\pdfoutput=1 

\documentclass[a4paper,10pt]{article}
\usepackage[utf8x]{inputenc}
\usepackage{amsmath,amssymb}
\usepackage{url}
\usepackage{tcolorbox}
\usepackage{xcolor, graphicx}
\usepackage[left=1.0in, right=1.0in, top=1.0in,bottom=1.0in]{geometry}
\usepackage{enumitem}
\setlist{nolistsep}
\usepackage{authblk}

\usepackage{fontawesome}
\usepackage{pifont}

\definecolor{qblue}{RGB}{26, 31, 113}
\definecolor{qred}{RGB}{139, 0, 0}
\usepackage[hidelinks]{hyperref}

\newcommand{\CSE}{CSE}

\newcounter{recno}
\newtcolorbox{essrec}[1][]{colframe=qblue, title={\refstepcounter{recno} \therecno~$\bullet$~#1}}

\title{Verification and Validation for Trustworthy Scientific Machine Learning}
\author[1]{John~D.~Jakeman} 
\author[2]{Lorena~A.~Barba}
\author[3]{Joaquim~R.~R.~A.~Martins}
\author[4]{Thomas~O'Leary-Roseberry}

\affil[1]{Optimization and Uncertainty Quantification, Sandia National Laboratories, Albuquerque, NM, 87123, USA}
\affil[2]{Department of Mechanical and Aerospace Engineering, The George Washington University, Washington, DC, 20052, USA}
\affil[3]{Department of Aerospace Engineering, University of Michigan, Ann Arbor, MI, 48109, USA}
\affil[4]{Oden Institute for Computational Engineering and Sciences, The University of Texas at Austin, TX, 78712, USA}

\date{}
\begin{document}
\maketitle

\begin{abstract}
Scientific machine learning (SciML) models are transforming many scientific disciplines. However, the development of good modeling practices to increase the trustworthiness of SciML has lagged behind its application, limiting its potential impact. The goal of this paper is to start a discussion on establishing consensus-based good practices for predictive SciML. We identify key challenges in applying existing computational science and engineering guidelines, such as verification and validation protocols, and provide recommendations to address these challenges. Our discussion focuses on predictive SciML, which uses machine learning models to learn, improve, and accelerate numerical simulations of physical systems. While centered on predictive applications, our 16 recommendations aim to help researchers conduct and document their modeling processes rigorously across all SciML domains.
\end{abstract}

\section{Introduction}
Scientific machine learning (SciML) integrates machine learning (ML) into scientific workflows to enhance system simulation and analysis, with an emphasis on computational modeling of physical systems.
This field emerged from Department of Energy workshops and initiatives starting in 2018, which also identified the need to increase ``the scale, rigor, robustness, and reliability of SciML necessary for routine use in science and engineering applications''~\cite{baker2019workshop}.
The field's subsequent growth through funding initiatives, conference themes, and high-profile publications stems from its ability to unite ML's predictive power with the domain knowledge and mathematical rigor of computational science and engineering (\CSE{}).
However, this surge in SciML development has outpaced good practices and reporting standards for building trust~\cite{Mcgreivy_H_arxiv_2024,Kapoor_et_al_SA_2024,Wang_MKetal_COM_2020,Zhu_YR_EST_2023,Xu_KBGW_RESS_2023}.

SciML models must demonstrate trustworthiness to be safe and useful~\cite{Jacovi_MMG_ACM_2021}.
Organizational and computational trust definitions~\cite{schneider1999trust,vashney2022trustworthy} inform our criteria for trustworthy SciML: competence in basic performance, reliability across conditions, transparency about processes and limitations, and alignment with scientific objectives.
These criteria span technical attributes (correctness, reliability, safety) and human-centric qualities (comprehensibility, transparency).\footnote{The notion of trustworthiness in SciML differs from ``trustworthy AI'' in socio-technical systems. While trustworthy AI addresses societal impacts, ethical considerations, and human behavioral aspects, trustworthiness in SciML focuses on mathematical rigor, physical consistency, and computational reliability in scientific and engineering applications.}
Trustworthy SciML performs consistently across operating conditions while providing insight into decision-making and adhering to scientific principles.
The model's decision process must align with prior information (intrinsic trustworthiness) and generalize to unseen cases (extrinsic trustworthiness)~\cite{Jacovi_MMG_ACM_2021}.

Rigorous modeling practices require sustained effort from the scientific community.
\CSE{} has established standards over decades for model reporting and assessment, including verification and validation, across biology~\cite{Patterson_W_CVM_1017}, earth sciences~\cite{Oreskes_SB_Science_1994, Jakeman_LN_EMS_2006}, and engineering~\cite{Schwer_EWC_2007, Roy_O_CMAME_2011, Sandkararaman_M_RESS_2015, AIAA_validation_report_1998}.
These standards help users and decision-makers document limitations, uncertainties, and modeling choices to prevent resource misallocation and maintain model credibility.
Similar SciML efforts are ``indispensable''~\cite{Qin_N_RESS_2023} but have only recently emerged in scientific disciplines, including materials science~\cite{Wang_MKetal_COM_2020}, environmental science~\cite{Zhu_YR_EST_2023}, nuclear engineering~\cite{Wang_LD_RESS_2024}, and aviation~\cite{Jia_YDJC_RESS_2023}, alongside more general initiatives~\cite{Kapoor_et_al_SA_2024}.
Developers and users of SciML must establish best practices for the development, reporting, and evaluation of new SciML paradigms.

This paper initiates a dialogue toward consensus-based practices for predictive SciML—the use of ML models to learn, improve, or accelerate physical system predictions. Rather than creating a rigid checklist, we guide transparent modeling processes by adapting existing \CSE{} verification and validation (V\&V) standards~\cite{AIAA_validation_report_1998,VV10-2006,Oberkampf_T_PAS_2002} while addressing SciML's unique challenges. Our guidance applies to both the development and deployment in support of a scientific claim and new algorithm development.

The remainder of the paper is structured as follows. Section~\ref{sec:background} defines our scope, compares \CSE{} and SciML development processes, and presents our four-component framework for SciML model development.
Sections~\ref{sec:problem-def}--\ref{sec:ongoing} detail these components and develop the recommendations.
The resulting 16 recommendations are listed below for quick reference.
Section~\ref{sec:conclusions} summarizes the key messages.

\begin{tcolorbox}[
    title=Recommendations for Trustworthy Scientific Machine Learning,
    colback=white,
    colframe=qblue,
    colbacktitle=qblue,
    coltitle=white,
    fonttitle=\bfseries
]

\noindent\textbf{Problem Definition}
\begin{enumerate}
\item Specify prior knowledge and model purpose
\item Specify verification, calibration, validation, and application domains
\item Carefully select and specify quantities of interest
\item Select and document model structure
\end{enumerate}

\noindent\textbf{Verification}
\begin{enumerate}[resume]
\item Verify code implementation with idealized test problems
\item Verify solution accuracy with realistic benchmarks
\end{enumerate}

\noindent\textbf{Validation}
\begin{enumerate}[resume]
\item Perform probabilistic calibration
\item Validate model against purpose-specific requirements
\item Quantify prediction uncertainties
\end{enumerate}

\noindent\textbf{Continuous Credibility Building}
\begin{enumerate}[resume]
\item Document data characteristics and impact
\item Document data processing procedures
\item Quantify SciML model sensitivities
\item Document the hyperparameter selection process
\item Use software testing and ensure reproducibility
\item Compare developed SciML model against alternatives
\item Explain the SciML prediction mechanism
\end{enumerate}
\end{tcolorbox}

\section{Background}
\label{sec:background}

While SciML is becoming increasingly used in \CSE{}, it differs fundamentally from \CSE{} in predicting system behavior. Traditional \CSE{} follows a deductive approach, deriving mathematical equations from conservation laws and physical properties to describe system causality.
Moreover, \CSE{} is used to predict the behavior of complex systems that are difficult or impossible to study experimentally~\cite{ruede2018research}.
In contrast, ML is an inductive discipline that learns relationships, potentially non-causal, directly from data or existing numerical models~\cite{Ruden_et_alAIDA_2020}.
SciML is a subset of ML and is an interdisciplinary field aimed at accelerating scientific and engineering discoveries.
While ML was once used only to help identify patterns in data from scientific instruments and sensors, SciML has evolved beyond data analysis to accelerate diverse scientific tasks, as discussed in the next section.

\subsection{Scope}
\label{sec:scope}

This paper focuses on SciML that enhances computational modeling and simulation of physical systems. Specifically, we address SciML approaches that support computational workflows where a model must be evaluated repeatedly within larger algorithmic procedures. In scientific computing, these are known as ``outer-loop'' processes because they form an additional computational layer that wraps around the core model solution. Such processes include:
(1) \emph{explanatory modeling}, which searches for insights into complex system behaviors; (2) \emph{uncertainty quantification}, which infers model parameters from observations and propagates them to predictive distributions~\cite{Ghanem_HO_book_2016}; (3) \emph{sensitivity analysis}, which identifies factors with greatest influence on predictions~\cite{Saltelli_book_2004}; (4) \emph{experimental design}, which predicts which data collections maximize understanding or decision value~\cite{Ryan_DMP_ISR_2016}; and (5) \emph{optimal design and control}, which optimizes objective functions by adjusting decision variables under constraints~\cite[Ch.~1]{Martins_Ning_book_2021}.
To support these outer-loop processes, predictive SciML encompasses four main classes of methods, each presenting different computational challenges: surrogate modeling, model discovery, hybrid \CSE{}-SciML modeling, and outer-loop learning.

\emph{Surrogate modeling} builds low-cost approximations of computationally expensive models for outer-loop analyses. 
Surrogate models can map model inputs to specific quantities of interest~\cite{Cohen_M_SMAIJCM_2017, Marrel_I_RESS_2024}.
Surrogate models can also approximate complete solutions to governing equations via data-driven reduced-order modeling~\cite{Peherstorfer_W_CMAME_2016, Schmid_JFM_2010}, operator learning with Gaussian processes~\cite{mora2024operator} and neural networks~\cite{Lu_JPZK_NMI_2021, Li_KBALBSA_icml_2021}, and physics-informed neural networks (PINNs)~\cite{yu2018deep,Raissi_PK_JCP_2019}, which encode physical constraints directly into the neural network training.
Finally, surrogate models can provide solutions to differential equations without simulation data~\cite{Sirignano_S_JCP_2018, W_B_CMS_2018}.

\emph{Model discovery} learns equations approximating physical system principles~\cite{Schmid_JFM_2010,Brunton_PK_PNAS_2016,Ling_JT_JCP_2016,Cory_CDEL_NC_2024}. Like operator learning and reduced-order modeling, these approaches work with both experimental and simulation data. Model discovery is being increasingly used to develop constitutive closure models that can be used when a numerical model cannot practically resolve small scales such that the large-scale equations are unclosed.

\emph{Hybrid \CSE{}-SciML models} embed SciML components within \CSE{} models.
Examples include learning initial guesses for iterative PDE solvers~\cite{huang2020int}, predicting preconditioners for linear systems~\cite{sappl2019deep}, and developing constitutive models for applications ranging from radiation transport~\cite{Huang_YCRY_MMS_2023} to aerospace~\cite{Singh_MD_AIAA_2017}.

\emph{Outer-loop learning} leverages ML to predict solutions under varying conditions.
Applications include surrogate models for expensive forward simulations in PDE-constrained optimization~\cite{wang2021fast} and Bayesian inverse problems~\cite{cao2024lazy}.
Reinforcement learning solves PDE-constrained optimization across different conditions~\cite{sutton2018reinforcement}.
ML methods can also learn solutions to these problems directly.
For example, ML models learn Bayesian posterior distributions as marginals of joint distributions for fast online inference with arbitrary observational data~\cite{baptista2020conditional,weilbach2020structured}.

\subsection{Comparing \CSE{} and SciML model development and deployment}

\begin{figure}[htb]
    \centering
    \includegraphics[width=\linewidth]{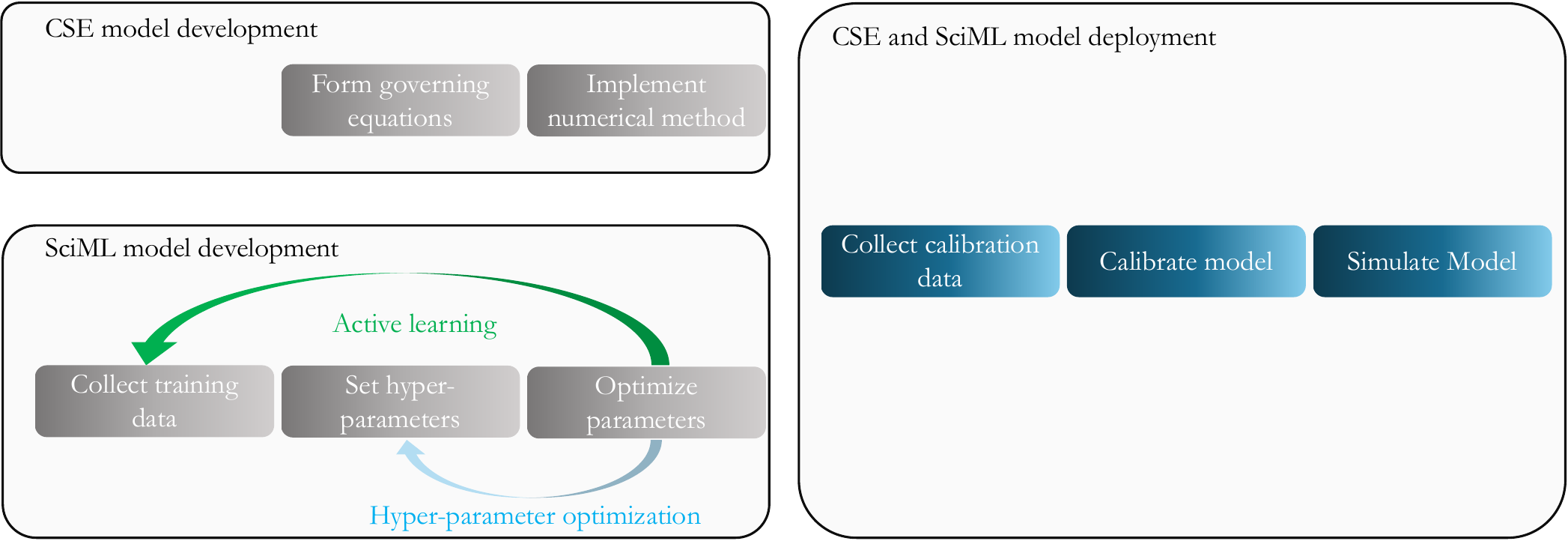}
    \caption{Fundamental steps for developing and deploying \CSE{} and SciML models in outer-loop processes. Deployment phases are conceptually similar for both model types when making scientific claims. Simulation steps require multiple model runs based on outer-loop process demands.}
    \label{fig:basic-sciml-workflow}
\end{figure}

Making scientific claims using SciML models involves two steps: model development and model deployment. 
While SciML models can complement \CSE{} models, their development processes differ, though deployment processes remain similar, as shown in Figure~\ref{fig:basic-sciml-workflow}.
In this paper we discuss the process of quantifying uncertainty in a hybrid SciML-\CSE{} ice-sheet model of the Humboldt Glacier in Greenland, with unknown glacier-landmass friction, to provide an example of how these processes differ.
Our example is loosely based on the study by He et al.~\cite{He_PKS_JCP_2023}, which exemplifies thorough reporting of a SciML model's strengths and weaknesses.

\CSE{} model development begins with physics-based governing equations. In our example, nonlinear Stokes equations accurately model shear-thinning behavior for ice sheets, but simpler models like shallow-shelf approximation (SSA) are often used for computational efficiency~\cite{He_PKS_JCP_2023}.
The next step implements numerical methods to solve the equations while respecting properties like mass and energy conservation, using techniques such as finite elements,  Newton iteration, and time-stepping to compute the coefficients of a fixed ansatz (basis) used to represent the solution. For instance, He and coworkers~\cite{He_PKS_JCP_2023} used continuous piecewise linear elements for thickness and velocity fields with Newton iteration and semi-implicit time-stepping.

In contrast, SciML typically begins by collecting input-output training data from either physical experiments or simulation. For example, He et al.~\cite{He_PKS_JCP_2023} built a DeepONet~\cite{Lu_JPZK_NMI_2021} to map ice-sheet thickness and friction to glacier velocity, using data from an SSA-based \CSE{} model. 
The second step sets model hyperparameters (e.g., neural-network width and depth). The subsequent steps involve optimizing the model parameters (e.g., neural-network weights and biases) to minimize the difference between model outputs and training data. He et al.~\cite{He_PKS_JCP_2023} used Adam optimization to minimize the mean-squared error between predicted and training velocities. 
Note that often, hyperparameters must be tuned through formal optimization or informal assessment of model accuracy across different settings. Additionally, some SciML models~\cite{Sirignano_S_JCP_2018, W_B_CMS_2018} combine \CSE{} and SciML development steps shown in Figure~\ref{fig:basic-sciml-workflow}, training without data by minimizing objectives that enforce known governing equations.

The development and deployment of SciML models for specific scientific applications follows distinct phases. When using these models to make scientific claims—--like predicting the Humboldt glacier's impact on sea level in the future—--deployment involves three key steps: 1) collecting calibration data (distinct from training data) representing the system's current state; 2) calibrating the model through optimization to match this system-specific data, which includes solving an inverse problem for friction parameters based on recent velocity measurements~\cite{Jakeman_SHHHP_ESD_2024}, notably optimizing model inputs rather than SciML parameters like neural-network weights; and 3) running the model under various future conditions or plausible friction fields, such as those obtained from Bayesian calibration.

\subsection{Leveraging an existing \CSE{} framework for trustworthy SciML}
\label{sec:framework}

As highlighted by Broderick et al.~\cite{Broderick_GMSZ_SA_2023}, trust in SciML and \CSE{} models can be gained or lost at any step of development and deployment (Figure~\ref{fig:basic-sciml-workflow}). Building trust requires: 1) demonstrating that the mathematical model adequately meets accuracy goals despite being an approximation; 2) verifying it by comparing theoretical and empirical results to ensure it solves its mathematical formulation correctly; 3) rigorously testing code to reduce software bugs; and 4) documenting data provenance to ensure relevance for future conclusions and decisions.

Over decades, the scientific computing community developed rigorous standards for building credibility in computational models. These standards crystallized into a systematic approach known as \emph{verification and validation} (V\&V), providing guidelines to enhance model trustworthiness by addressing potential failure points~\cite{AIAA_validation_report_1998,VV10-2006,Oberkampf_T_PAS_2002}. 
Verification ensures correct algorithm and code implementation (solving the equations correctly), while validation assesses model appropriateness for intended applications (solving the correct equations).

\begin{figure}[htb]
    \centering
    \includegraphics[width=\linewidth]{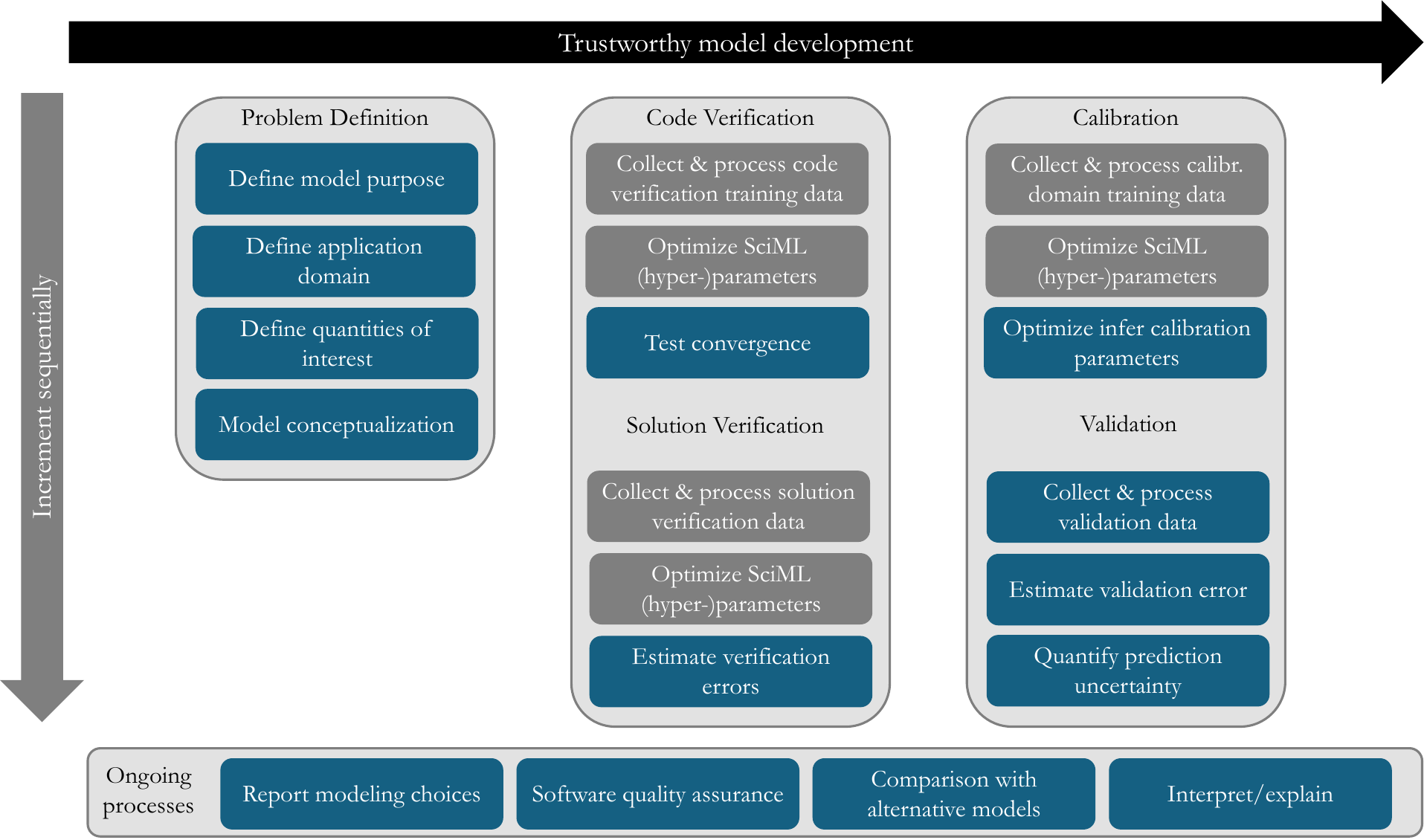}
    \caption{Four components of trustworthy model development. Blue boxes indicate areas common to \CSE{} and SciML models, while gray boxes show SciML-specific areas. Generative physics-informed SciML models may skip the gray data collection and processing boxes in code and solution verification.}
    \label{fig:model-development}
\end{figure}

While V\&V practices are well-established in scientific computing, the emergence of SciML brings new concerns that existing guidelines don't fully address. SciML combines aspects of both scientific computing and ML, presenting unique verification and validation challenges beyond either field. We argue that adapting proven V\&V principles can help provide a foundation for building trustworthy SciML models. 
To this end, we make recommendations in four areas: problem definition, verification, validation, and continuous credibility building (see Figure~\ref{fig:model-development}). 
Problem definition and verification (Sections~\ref{sec:problem-def} and~\ref{sec:verification}) expand on the model development steps in Figure~\ref{fig:basic-sciml-workflow},  while validation and reporting (Sections~\ref{sec:validation} and~\ref{sec:ongoing}) detail the deployment steps. Validation is generally most pertinent only to papers that utilize SciML to support a scientific claim for a specific application, but all other components illustrated in Figure~\ref{fig:basic-sciml-workflow} are applicable to all SciML papers, including those that present algorithmic advances relevant to a broad spectrum of scientific applications.

Like Broderick et al.~\cite{Broderick_GMSZ_SA_2023}, we view trustworthiness as a spectrum rather than binary. Following Oberkampf's predictive capability maturity model~\cite{oberkampf2007predictive}, the four components in Figure~\ref{fig:model-development} build confidence through continuous testing and refinement as new data emerge. Each study should mature progressively in its development process through these four areas. The following sections identify opportunities for improving SciML practice, along with corresponding recommendations. While, we do not claim that these four components or our recommendations are comprehensive or universally applicable, we aim to stimulate community efforts to develop SciML guidelines and standards for SciML development and deployment.
 
\section{Problem definition}
\label{sec:problem-def}

The first step in building and using a \CSE{} or SciML model is defining the problem scope: the model's intended purpose, application domain and operating environment, required quantities of interest (QoI) and their scales, and how prior knowledge informs model conceptualization.

\subsection{Model purpose}

\begin{essrec}[Specify prior knowledge and model purpose]
  Define the model's intended use and document essential properties, limitations, and constraints of the chosen approach. This ensures appropriate data selection and physics-informed objectives, preventing misuse outside the model's intended scope.
\end{essrec}

A SciML model's purpose, as discussed in Section~\ref{sec:scope}, dictates all subsequent modeling choices, including required outer-loop processes and essential properties.
For example, an explanatory model must simulate all system processes, such as ice-sheet thickness and velocity evolution, while a risk assessment model focuses only on decision-relevant quantities like mass loss under varying emissions scenarios.
In contrast, design and control models have different requirements, and the model's purpose influences the data types and formulations needed for training.
Therefore, the model formulation should be chosen based on these problem-specific considerations.

\subsection{Verification, calibration, validation and application domain}

\begin{essrec}[Specify verification, calibration, validation, and application domains]
  Define the specific conditions for the model's operation during verification, calibration, validation, and prediction phases, characterized by different boundary conditions, forcing functions, geometry, and timescales.
  Address potential differences between these domains and data distribution shifts to ensure the model architecture and training data align with intended use, preventing unreliable predictions outside validated conditions.
\end{essrec}

\begin{figure}[htb]
    \centering
    \includegraphics[width=0.65\linewidth]{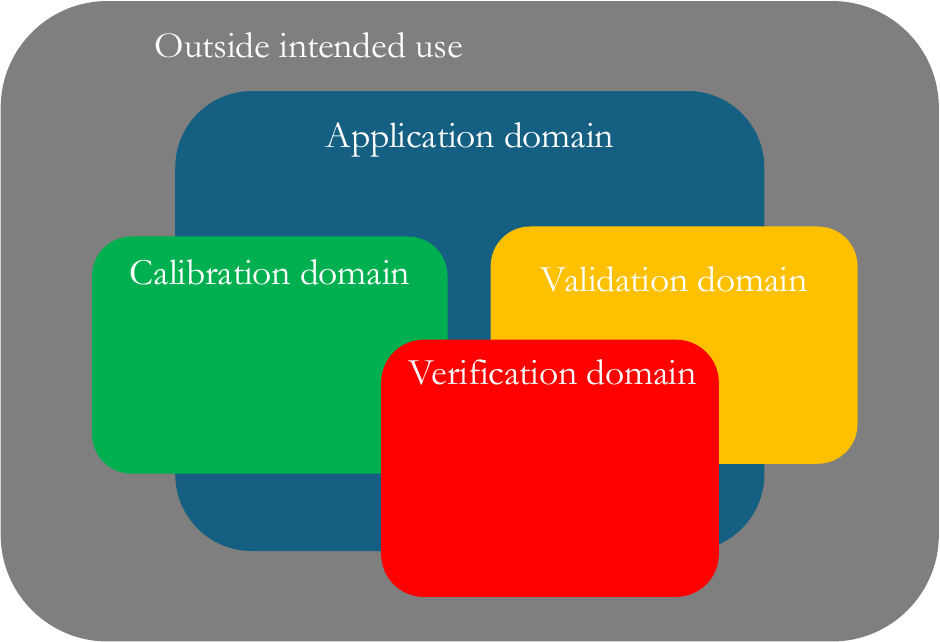}
    \caption{Verification, validation, calibration and application domains.}
    \label{fig:computational-domains}
\end{figure}

Trustworthy model development and deployment require assessing performance across various operational modes or domains~\cite{Chen_BD_RESS_2024}.
We advocate for evaluating models in four key domains: verification, calibration, validation, and application (see Figure~\ref{fig:computational-domains}), defined by the conditions under which the model operates, including boundary conditions, forcing functions, geometry, and timescales.
For ice sheets, examples include surface mass balance, land mass topography, and ocean temperatures.

Each domain typically requires predicting different quantities of interest under varying conditions, with process complexity increasing from verification to application, while the amount of supporting data decreases.
For instance, the verification domain predicts the entire state of the ice sheet using simple solutions, while the calibration domain focuses on Humboldt Glacier surface velocity under steady-state conditions, the validation domain predicts grounding-line change rates from the last decade, and the application domain forecasts glacier mass change in 2100.
Data shifts between domains must be considered; for example, a model trained solely on recent calibration data may struggle to predict ice-sheet properties under different future conditions.

\subsection{Quantities of interest}

\begin{essrec}[Carefully select and specify the quantities of interest]
Select and specify the model outputs (quantities of interest, QoI) needed for the intended use, identifying the minimal set for risk assessment and design applications, and a broader range for explanatory modeling.
  This choice influences the model's complexity, training data requirements, and computational approach for reliable predictions.
\end{essrec}

Quantities of interest (QoI) are the model outputs required by users, with their form and scale depending on the modeling purpose and application domain.
Risk assessment focuses on decision-critical QoIs, such as sea-level rise and infrastructure damage costs for ice sheets, while design applications require fewer QoIs to evaluate objectives and constraints, like thermal and structural stresses in aerospace vehicles.
In contrast, Design models need accurate QoI predictions only along optimizer trajectories while risk assessment models must predict across many conditions
\footnote{For any iteration of a risk averse design optimization the model may still need to be accurate across all uncertain model inputs}.
Whereas, explanatory modeling requires a broader range of QoIs, such as complete ice-sheet depth and velocity fields for studying calving, indicating that simple surrogates may suffice for risk assessment and design, but explanatory modeling may necessitate operators or reduced-order models.

\subsection{Model conceptualization}

\begin{essrec}[Select and document model structure]
  Choose a model structure that aligns with the model's purpose and domain, informed by relevant prior knowledge such as conservation laws.
  Document alternative structures considered and the reasoning behind the final selection, including the impact of resources and computational constraints, to ensure a balance of usability, reliability, and transparency about assumptions and limitations.

\end{essrec}

Model conceptualization, following problem definition, involves selecting model structure based on prior knowledge and requires clear identification of the application domain and relevant QoI. Key prior knowledge, such as conservation laws and system invariances, should guide method selection -- for example, symplectic time integrators can be used to preserve system dynamics~\cite{ruth1983canonical}.

A \CSE{} modeler chooses between model types, such as lumped versus distributed PDE models, and linear versus nonlinear PDEs, with the optimal depending on the application domain, QoI, and available resources. For example, linear PDEs may introduce more error but their lower computational cost enables better error and uncertainty characterization for tasks like optimal design.
Similar considerations guide SciML model selection. For example, Gaussian processes excel at predicting scalar QoI with few inputs and limited data, but become intractable for larger datasets without variational inference approximations~\cite{Liu_CO_KBS_2018}. In contrast, deep neural networks handle high-dimensional data but require large datasets. The intended use also shapes model structure and training, e.g., optimization applications require controlling derivative errors~\cite{bouhlel2020scalable} to ensure convergence~\cite{luo2023efficient}.

\CSE{} has a strong history of using prior knowledge to formulate governing equations for complex phenomena and deriving numerical methods that respect important physical properties. However, all models are approximate and the best model must balance usability, reliability, and feasibility~\cite{Hamilton_PSFJEMS_2022}. While SciML methods can be usable and feasible, more attention is needed to establish their trustworthiness, which we will address in the following two sections on V\&V.

\section{Verification}
\label{sec:verification}

Verification increases the trustworthiness of numerical models by demonstrating that the numerical method can adequately solve the equations of the desired mathematical model and the code correctly implements the algorithm. Verification consists of code verification and solution verification, which enhance credibility and trust in the model's predictions. Code and solution verification are well-established in \CSE{} to reduce algorithmic errors. However, verification for SciML models has received less attention due to the field's young age and unique challenges. Moreover, because SciML models heavily rely on data, unlike \CSE{} models, existing \CSE{} verification notions need to be adapted for SciML.

\subsection{Code verification}
\label{sec:code-verification}

\begin{essrec}[Verify code implementation with test problems]
  Evaluate the SciML model's accuracy on simple manufactured test problems using verification data independent from the training data, assessing how model error responds to variations in training samples and optimization parameters while increasing model complexity and training data size.
  This systematic approach reveals implementation issues, quantifies the impact of sampling and optimization choices, and builds confidence in the model's numerical implementation.
\end{essrec}

Code verification ensures that a computer code correctly implements the intended mathematical model.
For \CSE{} models, this involves confirming that numerical methods and algorithms are free from programming errors (``bugs").
PDE-based \CSE{} models commonly use the method of manufactured solutions (MMS) to verify code against known theoretical convergence rates by substituting a user-provided solution into the governing equations.
If the observed order of convergence is less than theoretical, potential causes such as software bugs or insufficient mesh refinement must be identified.

Code verification for SciML models is also important but challenging due to the large role of data and nonconvex numerical optimization.
Three main challenges limit code verification for many SciML models:
while theoretical analysis of SciML models is increasing~\cite{schwab2023deep,opschoor2022exponential, lanthaler2023curse,kovachki2023neural}, many models like neural networks do not generally admit known convergence rates outside specific map classes~\cite{schwab2023deep,opschoor2022exponential,herrmann2024neural};
there are not general procedures for refining models, e.g. changing neural-network width and depth as data increases, 
and generalization error often plateaus due to nonconvex optimization issues like local minima and saddle points~\cite{Dauphin_PGCGB_NIPS_2014}.

Addressing these challenges through theoretical and algorithmic advances can enhance the trustworthiness of SciML models.
Currently, convergence-based verification is feasible for certain models with established theory, such as operator methods~\cite{Turnage_et_al_arxiv_2024}, polynomial chaos expansions~\cite{Cohen_M_SMAIJCM_2017}, and Gaussian processes.
For models without supporting theory, convergence tests should still be conducted and reported, as evidence of model convergence increases trustworthiness.
For example, observing Monte Carlo-type sampling rates in a regime of interest for a fixed overparametrized model can provide intuition into whether the model should be enhanced.

To adapt code verification for SciML models, first, report errors for varying training data realizations to quantify sampling error. Second, conduct verification studies using artificially generated data from a random realization of a SciML model, comparing recovered parameter values or predictions with true values. Additionally, quantify the sensitivity of model error to randomness in optimization by varying the random seed and initial guess (see Section~\ref{sec:loss-and-opt}). All verification tests must use independent test or \emph{verification data} to measure the accuracy of the SciML model.

\subsection{Solution verification}

\begin{essrec}[Verify solution accuracy with realistic benchmarks]
  Test the model's performance on realistic benchmark problems that reflect the intended application domain, quantifying how model error varies with different training data samples and optimization parameters.
  When possible, examine error patterns across various model complexities and data amounts; otherwise, focus on verifying the specific configuration for deployment to ensure accuracy under realistic conditions.
\end{essrec}

Code verification establishes a code's ability to reproduce known idealized solutions, while solution verification, performed after code verification, assesses the code's accuracy on more complex yet tractable problems defined by more realistic boundary conditions, forcing, and data.
For example, ice sheets code verification may use manufactured solutions, whereas solution verification may use more realistic MISMIP benchmarks~\cite{Cornford_et_al_TC_2020}.
In solution verification, the numerical solution cannot be compared to a known exact solution and convergence rate, thus other procedures must be used to estimate the error introduced by numerical discretization.

Solution verification establishes whether the exact conditions of a model result in the expected theoretical convergence rate or if unexpected features like shocks or singularities prevent it.
The most common approach for \CSE{} models compares the difference between consecutive solutions as the numerical discretization is refined and uses Richardson extrapolation to estimate errors.
A posteriori error estimation techniques that require solving an adjoint equation can also be used.

While thorough solution verification of CSE models is challenging, it is even more difficult for SciML models.
Currently, solution verification of SciML models simply consists of evaluating a trained model's performance on test data separate from the training data, which is insufficient as it does not quantify the impact of increasing data and model complexity on model error.
A posteriori error estimation using techniques like Richardson extrapolation is complicated by model expressivity, statistical sampling errors, and variability from converging to local solutions in nonconvex optimization.

Until convergence theory for SciML models improves and automated procedures for adjusting hyperparameters as data increases are developed, solution verification should repeat the sensitivity tests proposed for code verification (Section~\ref{sec:code-verification}), with two key differences:
First, verification experiments must be specifically designed for solution verification, as not all verification data equally informs solution verification efforts, similar to observations made when creating validation datasets for \CSE{} models~\cite{Oberkampf_T_PAS_2002} and discussed further in Section~\ref{sec:data-sources}.
Second, while investigating the convergence of SciML errors on realistic benchmarks is ideal, it may be computationally impractical; thus, solution verification should prioritize quantifying errors based on the model complexity and data amount intended for deployment.

\section{Validation}
\label{sec:validation}

Verification establishes if a model can accurately produce the behavior of a system described by governing equations. In contrast, validation assesses whether a \CSE{} model's governing equations---or data for SciML models---and the model's implementation can reproduce the physical system's important properties, as determined by the model's purpose.

Validation requires three main steps: (1) solve an inverse problem to calibrate the model to observational data; (2) compare the model's output with observational data collected explicitly for validation; and (3) quantify the uncertainty in model predictions when interpolating or extrapolating from the validation domain to the application domain. We will expand on these steps below.
But first note that the issues affecting the verification of SciML models also affect calibration and validation. Consequently, we will not revisit them here but rather will highlight the unique challenges in validating SciML models.

\subsection{Calibration}

\begin{essrec}[Perform probabilistic calibration]
  Calibrate the trained SciML model with observational data to optimize predictive accuracy, using Bayesian inference when possible to generate probabilistic parameter estimates and quantify model uncertainty.
  Select calibration metrics that account for both model and experimental uncertainties, and strategically choose calibration data to maximize information content, ensuring reliable uncertainty estimation.
\end{essrec}

Once a \CSE{} model has been verified, it must be calibrated to match experimental data that contains observational noise, which involves solving either a deterministic or statistical (e.g., Bayesian) inverse problem~\cite{Stuart_AN_2010}.
The deterministic approach formulates this as a (nonlinear) optimization problem that minimizes the mismatch between model and experimental data, requiring regularization methods like the  L-curve~\cite{hansen1999curve} or the Morozov discrepancy principle~\cite{anzengruber2009morozov}.
In contrast, the Bayesian approach replaces the misfit with a likelihood function, based on the noise model, while using a prior distribution for regularization. We recommend Bayesian methods for calibration because they provide insight into the uncertainty of the reconstructed model parameters. 

Calibrating SciML and hybrid CSE-SciML models is distinct from SciML training and follows similar principles to \CSE{} model calibration.
These models are first trained using simulation data for solution verification.
Next, observational data (called \emph{calibration data}) determines the optimal model input values that match experimental outputs.
For instance, calibrating a SciML ice-sheet model, e.g. \cite{He_PKS_JCP_2023}, requires finding optimal friction field parameters of a trained SciML model, which best predict observed glacier surface velocities, accounting for observational noise.

While, calibration typically improves a model's predictive accuracy on its application domain, the informative value of calibration data varies significantly.
Therefore, researchers should select calibration data strategically to maximize information content within their experimental budget. See Section~\ref{sec:data-sources} for further discussion on collecting informative data.

\subsection{Model validation}

\begin{essrec}[Validate model against purpose-specific requirements]
  Define validation metrics aligned with the model's intended purpose and validate it using independent data not used for training to ensure it captures essential physics.
  If performance is inadequate, iterate by collecting more training data or refining the model until satisfactory accuracy is achieved, ensuring stakeholder requirements are met.  
\end{essrec}

Model validation is the ``substantiation that a model within its domain of applicability possesses a satisfactory range of accuracy consistent with the intended application of the model''~\cite{Refsgaard_H_AWR_2004}.
It involves comparing computational results with independent observational data, to determine if the agreement meets the model's intended purpose~\cite{Lee_et_all_AIAA_2016}.
For \CSE{} models with unacceptable validation agreement, modelers must either collect additional calibration data or refine the model structure until reaching acceptable accuracy, while SciML modelers can also collect more training data.

Validation must occur after calibration and requires independent data not used for calibration or training and well chosen validation criteria~\cite{Marrel_I_RESS_2024}.
In our conceptual ice-sheet model, calibration matches surface velocities representing pre-industrial conditions, while validation assesses the calibrated model's ability to predict grounding line change rates at the start of this decade.
Performance metrics must target the specific modeling purposes, such as measuring distance from true optima for optimization tasks~\cite{cao2024lazy} or quantifying errors in uncertainty statistics.
For explanatory SciML modeling, validation metrics must also assess physical fidelity, including adherence to physical laws and conservation properties.
As with verification, the validation concept should encompass \emph{data validation}, particularly whether training data adequately represents the application space.

Note that the term validation, as used here, differs from the concept of \emph{cross validation}, which estimates ML model accuracy on data representative of the training domain during development.
Validation determines whether a model is acceptable for its specific purpose rather than universally correct.
The definition of acceptable is subjective, depending on validation metrics and accuracy requirements established by model stakeholders in alignment with the problem definition and model purpose (see Section~\ref{sec:problem-def}).
The amount of data needed to meet these criteria will depend on the information content of the calibration data, which can vary substantially if not chosen judiciously.
Validation experiments should ``capture the essential physics of interest, including all relevant physical modeling data and initial and boundary conditions required by the code''~\cite{Oberkampf_T_NED_2008}.
Lastly, validation itself does not constitute final model acceptance, which must be based on model accuracy in the application domain, as discussed in Section~\ref{sec:prediction}.

\subsection{Prediction}
\label{sec:prediction}

\begin{essrec}[Quantify prediction uncertainties]
  Quantify all sources of uncertainty affecting model predictions in the application domain, such as numerical errors and sampling errors, and propagate these uncertainties to estimate relevant statistics that meet validation criteria.
  Set acceptance thresholds for prediction uncertainty to ensure model reliability while acknowledging the limitations of uncertainty quantification.
\end{essrec}

Although extensive data may be available for model calibration, validation data is typically scarcer and may not represent the model's intended application domain. According to Schwer~\cite{Schwer_EWC_2007}, ``The original reason for developing a model was to make predictions for applications of the model where no experimental data could, or would, be obtained.'' Therefore, minimizing validation metrics at nominal conditions cannot sufficiently validate a model. Modelers must also quantify accuracy and uncertainty when predictions are extrapolated to the application domain.

SciML models, like \CSE{} models, are subject to numerous sources of uncertainty that must be quantified, including: numerical errors, which arise from approximating the solution to governing equations; input uncertainty, which is caused by inexact knowledge of model inputs; parameter uncertainty, which stems from inexact knowledge of model coefficients; and model structure error representing the difference between the model and reality.
However, SciML models are also subject to sampling and optimization errors; sampling error arises from training with finite, potentially noisy data, while optimization error reflects the difference between local and global solutions, affecting SciML models during both calibration and training.
Linear approximations, for example, based on polynomials, achieve zero optimization error during training to machine precision.
However, nonlinear approximations such as neural networks often produce non-trivial optimization errors.
Stochastic gradient descent demonstrates this by producing different parameter estimates due to stochastic optimization randomness and initial guesses.

All uncertainties must be propagated onto predictions made in the application domain. Quantifying a single error, such as those reported by generative models like variational autoencoders or Gaussian processes, is typically insufficient.
All sources of uncertainty should be parameterized using expert knowledge to construct prior distributions, which are then updated through Bayesian inference and calibration data to form posterior distributions.
Methods like Monte Carlo quadrature can randomly sample from these posterior distributions to compute empirical estimates of important statistics defined by validation criteria, such as mean and variance.

While Monte Carlo-based UQ procedures effectively quantify parameterized uncertainties, model structure error remains challenging to parameterize, and validation can only partially assess it, as experiments rarely cover all conditions of use.
Progress has been made in quantifying extrapolation error for hybrid models that combine reliable physics-based equations with less-reliable embedded models~\cite{Oliver_TSM_CMAME_2015}, but pure SciML models still require significant research to develop reliable methods for estimating model structure uncertainty.
Moreover, complete elimination of uncertainty is impossible.
Consequently, model acceptance, like validation, must rely on subjective accuracy criteria established through stakeholder communication.
For example, acceptance criteria for predicted sea-level change from melting ice sheets by 2100 may specify a prediction precision of 1\% of the mean value, while engineering applications, such as aerospace design, may demand much higher accuracy requirements.

\section{Continuous Credibility Building}
\label{sec:ongoing}

The three model development components depicted in the top of Figure~\ref{fig:model-development} appear sequential. 
However, practical SciML development is iterative, often requiring steps to be repeated.
Poor validation metrics may necessitate model updates or additional data collection. 
Sometimes, simplifying objectives or the QoI scale may be warranted to achieve more accurate results for a simpler but still valuable problem. 

The iterative nature of modeling requires ongoing documentation of the critical aspects that affect V\&V including data sources and processing, loss functions and optimizers and model tuning. Additionally, it is important to perform software quality assurance (SQA), model comparisons, and explain and interpret predictions and model behaviors. Enacting these steps is referred to as continuous credibility building in Figure \ref{fig:model-development}.

\subsection{Data Sources}
\label{sec:data-sources}

\begin{essrec}[Document data characteristics and impact]
  Document the properties of all data sources, including fidelity, biases, and limitations, and explain their impact on model performance and verification processes.
 Assessing data requirements based on specific accuracy needs, quantify the and the impact of using lower-fidelity data will enhance reproducibility and transparency regarding the model's limitations.
\end{essrec}

The types of data used to train a SciML model significantly affect V\&V, so data sources must be carefully selected and documented.
Key properties to consider include data fidelity, bias, noise, and the cost of data collection, which can limit the amount available.
While some studies recommend data quantity based on the number of input features~\cite{Zhu_YR_EST_2023}, we believe it should be determined by the accuracy requirements of the model's intended use.
Future work on generalization accuracy and statistical learning theory for SciML models will help better elucidate these mathematical trade-offs.
Additionally, the fidelity of training data impacts model performance; for instance, using lower-fidelity simulation data may allow for more training samples but introduces bias relative to high-fidelity models.
This error's impact on predicted outcomes, such as the final objective value of an optimal design, must be quantified.
Alternatively, multi-fidelity SciML models~\cite{Cutajar_PDLG_Arxiv_2019} can mitigate this issue by combining limited high-fidelity data with larger amounts of low-fidelity data.

Creating informative benchmarks can significantly enhance the quality of V\&V analyses and facilitate comparisons of methods and SciML structures. However, unlike traditional ML, SciML and CSE lack commonly agreed-upon benchmark problems for solution verification and validation, necessitating further work to build on initial efforts to establish SciML benchmarks~\cite{Takamoto_PLMAPN_NIPS_2022}. Following Oberkampf's guidance~\cite{Oberkampf_TH_2004}, benchmarks should clearly state their purpose, comparison methods, and specific acceptance criteria; for instance, a SciML surrogate for uncertainty quantification should accurately compute the mean and variance of the true model to a specified level of accuracy, while a SciML model for optimization should compare its optimal solution to trusted reference optima. Additionally, benchmarks must be widely accessible, though this can be challenging with large datasets. Until community benchmarks and dissemination mechanisms are developed, researchers should thoroughly document their choices and explain the significance of the problems they address.

\subsection{Data processing}

\begin{essrec}[Document data processing procedures]
  Document procedures for handling missing or abnormal data and any transformations applied, along with the rationale and potential impact on model performance.
  This enhances reproducibility and helps identify potential sources of bias in preprocessing.
\end{essrec}

Before an ML model can be constructed, the data that will be used to train the ML model must be processed for missing and abnormal data.\footnote{It goes without saying that unprocessed raw data should be preserved and properly archived before any transformations are applied.} 
When building ML models from simulation data, missing data can occur when one or more simulations fail to complete, caused for instance by the model becoming unstable for a certain set of inputs, or the computer system crashing. Similarly, abnormal data can be generated when the instability of the model does not cause the simulation to fail but leads to errors, or when a soft fault occurs on a high-performance computing system.
The methods employed for dealing with abnormal data---such as removal, interpolation or imputation---must be defined as they can substantially impact the accuracy and credibility of a SciML model. 

Variable transformations are also often used to process data when training a SciML model. For instance, it is common to normalize the data to have zero mean and unit standard deviation, or to take the logarithm of the data with the goal of improving the predictive performance of a SciML model.
Modelers should report the use of any transformations because the use of different transformations can lead to substantial changes to the predictive performance of the ML model. 
Moreover, data processing scripts should be preserved in a trusted repository and published alongside other study artifacts.

\subsection{Loss functions and optimizers}\label{sec:loss-and-opt}

\begin{essrec}[Quantify SciML model sensitivity]
  Test and document the model's sensitivity to optimization parameters, including random initialization and regularization choices, while reporting all configurations and hyperparameters.
  This assessment quantifies the impact of variations on model performance, identifies potential instabilities, and ensures reproducibility.
\end{essrec}

Training an ML model requires minimizing a loss function that quantifies how well the model approximates the training data. This loss function is often the mean squared error or mean absolute error. Nevertheless, a plethora of other loss functions can be used. Indeed, the loss function should be tailored to the intended use of the ML model. For example, the mean squared error captures the average size of the squared residuals between the ML model and the data, but treats positive and negative residuals equally. Thus, if learning a function intended to capture large events it may be more appropriate to train the SciML model using asymmetric loss functions such as the quantile loss function~\cite{Jakeman_KH_RESS_2021}.

Additionally, the optimization method used to minimize the loss function can also substantially impact model trustworthiness. For example, it is common to use stochastic optimization methods such as Adam~\cite{Kingma_B_arxiv_2017} to train a SciML model. Thus, the optimal solution depends on the random seed, the initial condition and the computing platform used to perform the optimization. Consequently, the impact of this randomness must be quantified, for example, by reporting the variability in the error of the SciML model learned using different perturbations of these factors.

\subsection{Model tuning}

\begin{essrec}[Document the hyperparameter selection process]
  Document hyperparameter selection details, including the optimization method, value ranges, criteria for selection, and procedures, such as cross-validation, used to balance model complexity with performance.
  This documentation enhances reproducibility and interpretability, helping researchers understand the trade-offs in model development and ensuring the configuration is suitable for its intended application.
\end{essrec}

Training a SciML model is typically an iterative process, which involves optimizing the parameters of a model, e.g., weights and biases of a neural network, for different values of the model hyper-parameters, e.g., the depth and width of a neural network, also called hyperparameter tuning~\cite{Feurer_H_AMLMSC_2019}. 
Model structures with large numbers of parameters (common in SciML) can lead to overfitting, where the model performs well on the training data but poorly on new, unseen data. Additionally, an overly complex model can become less interpretable, making it harder for researchers to understand and explain the model's predictions. Furthermore, it typically becomes harder to identify the impact of each parameter, a problem known as reduced parameter identifiability~\cite{Guillaume_Jetal_EMS_2019}, as the number of model parameters increases. 
The final choice of hyperparameters can significantly impact a model's predictive power, interpretability, and the uncertainty of its predictions. Yet, despite the importance of hyperparameter tuning, a review of over 140 highly cited papers using ML for environmental research~\cite{Zhu_YR_EST_2023} found that only 19.6\% of those papers conducted formal hyperparameter optimization methods (such as grid search) and another 25.7\% used trial and error. To ensure that the scientific community can fully evaluate and understand published works in SciML, detailed documentation of these processes is essential. 

\subsection{Software quality assurance and reproducibility}

\begin{essrec}[Use software testing and ensure reproducibility]
  Implement comprehensive testing throughout the model development workflow, addressing ML-specific challenges such as stochastic behavior and emergent properties.
  Release complete code repositories that include source code, trained models, data, reproduction scripts, and detailed environment documentation to facilitate independent verification of results, support long-term maintenance, and build trust in the model's implementation.
\end{essrec}

Software quality assurance (SQA) is required by both \CSE{} and SciML models to minimize the impact of programming errors in model predictions, facilitate collaboration, and build credibility into the modeling activity. Some \CSE{} guidelines call for SQA as part of code verification~\cite{Oberkampf_T_PAS_2002}, given its focus on correctness of the software by finding and fixing programming errors or `bugs' that are prevalent in all software~\cite{Hatton_book_1997, Zhang_IEEECSS_2009}. 
While SQA is an essential component of code verification, we discuss it separately here to emphasize its importance to the entirety of \CSE{} and SciML workflows. 

SQA requires both static analysis~\cite{Louridas_IEEES_2006} and dynamic testing~\cite{Jamil_ANA_ICT4M_2016}.  
Static analysis checks for errors without executing code and can be applied to an entire code base before dynamic tests are executed.
This is useful because runtime tools only test the code that is executed and thus can miss corner cases that are not tested~\cite{Gulabovska_P_IEEE_2019}.
Dynamic testing comprises unit tests, integration tests and system tests.
Despite the importance of SQA, all of these software engineering practices have low adoption in ML~\cite{Seban_BHV_ACM_2020} and CSE  software~\cite{Kanewala_B_IST_2014}.
Software engineering is particularly challenging in SciML models because it has ``\ldots all of the problems of non-ML software systems plus an additional set of ML-specific issues''~\cite{Wan_IEEE_XLM_2021}.
While deterministic CSE can often use theoretically derived test ``oracles'' that capture the expected behavior of a model~\cite{Johanson_H_CSE_2018}, ML's stochastic nature requires aggregating multiple model runs to assess variability.
System-wide testing is often needed to understand emergent properties, making unit testing less useful.
Additionally, code coverage metrics are less meaningful for SciML since decision logic is relearned with each training~\cite{Zhang_HML_IEEE_TSE_2022}.
Improving SQA adoption is thus crucial for building trustworthy SciML models.
SciML studies should make every effort to establish confidence in the software implementation, via modular tests and benchmarks, and adoption of software engineering best practices. 

Reproducibility is essential for building the credibility of computational models~\cite{nasem_2019} and serves as a form of quality assurance, which is often missing from ML research~\cite{hutson2018reproducibility,haibe2020transparency}.
Rigorous documentation of the training pipeline is necessary, including details of the optimization and regularization applied, activation functions, hyperparameters, learning-rate schedules, data specifications, computing requirements, software versions, and public code repositories.
Repositories must be well-organized with tutorials and reproduction scripts~\cite{Kapoor_et_al_SA_2024}, store the optimized parameters and hyperparameters of SciML models, such as neural network architecture and weights, and standard public licenses should state reuse rights.
Simply depositing source code in a public repository is insufficient. Data provenance must be strictly managed, with full datasets ideally deposited in trusted repositories that offer persistent identifiers.
While these efforts may seem burdensome, they are crucial for enhancing SciML's trustworthiness and are increasingly emphasized by funders, journals, and conferences, ultimately benefiting authors by lowering barriers to the adoption of their methods.

\subsection{Cost-accuracy comparison with alternatives}

\begin{essrec}[Compare developed SciML model against alternatives]
  Compare the SciML model with alternative SciML approaches and traditional CSE methods using purpose-specific validation metrics, and document all computational costs, including data generation, parameter optimization, training, and inference times.
  Clearly report both limitations and negative results alongside successes, providing quantitative evidence to facilitate informed model selection and foster trust through a balanced presentation of strengths and weaknesses.
\end{essrec}

SciML models must be compared with state-of-the-art alternatives throughout model development, including verification, validation, and prediction stages. However, such comparisons are often neglected in literature~\cite{Saraygord_EXL_RESS_2022}. For example, a review of environmental SciML studies found that 37.8\% made no ML comparisons, 44.6\% compared with one SciML method, and only 14.9\% compared with multiple methods~\cite{Zhu_YR_EST_2023}.

Model comparison credibility depends on the types of SciML models compared. When a parsimonious model outperforms more complex ones~\cite{Tiantian_etal_WR_2017}, comparing it only within its model family is insufficient. Comparing a neural network with Gaussian processes and linear regression better contextualizes performance than comparing multiple neural networks. While comparing multiple ML methods is valuable, when CSE alternatives exist, SciML methods must also be evaluated against state-of-the-art CSE approaches using purpose-specific metrics, particularly those used in validation.

Although accuracy is crucial in method comparison, computational costs must also be documented, yet total modeling costs are often underreported. Researchers should detail all costs, including data generation (e.g., running a \CSE{} model), parameter optimization, and prediction.
For instance, in PDE-constrained optimization, direct PDE optimization costs should be compared to combined surrogate training and optimization costs. Additionally, dismissing training costs as ``offline" is misleading, however, these costs can be amortized when SciML models are used across multiple studies, as in learning to optimize~\cite{chen2022learning}. Nevertheless, such amortization claims need qualification since SciML models' cost-effectiveness---the trade-off between accuracy, development time, and cost---is harder to predict than for \CSE{} models~\cite{Sculley_etal_Neurips_2015}.

Finally, efforts must be made to avoid the misuse of language, as it can distort the interpretation of study findings and lead to "spin" or reporting bias~\cite{Constanza_et_al_JCE_2023}. For example, McGreivy and Hakim ~\cite{Mcgreivy_H_arxiv_2024} found that 79\% of 76 ML papers on fluid PDEs used weak baselines to claim superiority over standard methods, and of 232 abstracts surveyed 94.8\% reported only positive results, indicating widespread reporting bias.

\subsection{Explain and Interpret}

\begin{essrec}[Explain the SciML prediction mechanism]
  Explain the model's predictions by detailing its structural features and decision-making processes, specifying any enhanced aspects of interpretability (e.g., parsimony, robustness) with quantitative evidence.
  This clarification builds trust in the model's predictions and allows domain experts to validate its behavior against scientific understanding.
\end{essrec}

SciML and \CSE{} researchers should explicitly identify how their methods advance specific scientific objectives. Sufficient information should be provided ``to infer the purpose of the technology and set their expectations sensibly based on such understanding''~\cite{Toreini_ACEZM_ACM_2020}.
Model trustworthiness depends both on the process and its results.
Consequently, researchers should explain how they came to their conclusions.
According to Doshi-Velez and Kim~\cite{Doshi_K_arxiv_2017}, this can involve experts interpreting highly intelligible models, where interpretability is built into their model structure. 
For example, a physicist might understand the reasons for the bifurcation of a dynamical system or use a second model to understand the learned model's behavior. 
Doshi-Velez and Kim~\cite{Doshi_K_arxiv_2017} also advocate for evaluating algorithm robustness to parameter and input variations and for validating that predicted changes in model outputs correspond to real system behavior.

The limited interpretability and transparency of machine learning (ML) models pose significant challenges in critical fields like medicine and aerospace engineering. The complexity of Scientific Machine Learning (SciML) models can hinder trust in their predictions, particularly when decisions impact lives. Unfortunately, a universal definition of interpretability remains elusive due to its subjective nature and variation across disciplines~\cite{Lipton_arxiv_2017}. Linear models are often deemed interpretable because their coefficients indicate feature importance; however, this clarity diminishes with thousands of parameters~\cite{Molnar_CB_chapter_2020}. Thus, claims of superior interpretability require thorough justification, presenting an opportunity for research communities to establish domain-specific criteria for interpretability, alongside explainability methods that clarify model predictions.

\section{Concluding remarks}
\label{sec:conclusions}
Scientific machine learning (SciML) is an emerging field that integrates machine learning into scientific workflows, creating a powerful synergy between traditional and new methodologies.
More specifically, we define SciML as the application of machine learning methods to enhance, accelerate, or improve the computational modeling and simulation of physical systems.
This definition is applicable across a wide array of scientific disciplines, including physics, earth sciences, and engineering.
However, we acknowledge that this definition may not be universally embraced in certain fields, such as psychology and social sciences.

While computational science and engineering (CSE) has developed solid theoretical foundations over decades, SciML's rapid empirical advances have often outpaced formal theoretical analysis.
Successfully advancing SciML requires thoughtfully integrating the complementary strengths of both fields: CSE's mathematical rigor and ML's data-driven innovations.
Moreover, the CSE community has established rigorous guidelines for model development, verification, and validation (V\&V) to support scientific claims.
In contrast, such standards are still emerging in SciML.
Consequently, 
this paper addresses the critical need for developing V\&V  practices for SciML that builds upon CSE frameworks while addressing SciML's unique challenges.

In this paper, we identified key challenges to developing trusted SciML practices, including the field's heavy reliance on data, the optimization of non-convex functions, and gaps in theoretical understanding.
To address these challenges, we presented a four-component framework for developing and deploying trustworthy SciML models.
As part of this framework, we established 16 recommendations to provide concrete guidance on critical aspects like reporting computational requirements and preparing detailed documentation. We also used real-world examples to illustrate concepts and offer actionable guidance for both development and deployment phases.

The recommendations in this paper are particularly crucial for high-consequence applications where model credibility is paramount. 
These include safety-critical systems in aerospace and nuclear engineering, digital twins for infrastructure monitoring, and medical applications that use model predictions guide treatment decisions. In such domains, the cost of model failure---whether from inadequate verification, poor validation, or incomplete uncertainty quantification---can be severe. Our framework provides a foundation for building the necessary trust in SciML models for these critical applications.

Beyond these domain-specific challenges, the rapid evolution of the field and the drive to publish quickly have led to poor reproducibility, often compromising rigor and transparency.
Here, we aim to catalyze broader community dialogue around establishing consensus-based practices for predictive SciML.
As such, this paper makes a meaningful contribution by being one of the first comprehensive attempts to address systematic V\&V practices specifically for SciML.
It fills an important gap in the literature, as while both CSE and ML have their own established practices, the intersection of these fields requires new approaches to ensure trustworthiness.

We encourage journal editors to adopt these, or similar, recommendations in their submission guidelines, funding agencies to require V\&V plans in proposals, conference organizers to promote sessions on SciML credibility, and researchers to implement these practices in their work. 
Only through collective effort can we establish rigorous standards needed to fully realize SciML's potential for accelerating scientific discovery and enabling trustworthy predictions in high-consequence applications.

\section*{Acknowledgments}
Funding: John Jakeman's work was funded by Sandia National Laboratories’ Laboratory Directed Research and Development (LDRD) program.
Thomas O'Leary-Roseberry's work was funded by the National Science Foundation under DMS Award 2324643 and OAC Award 2313033.
Sandia National Laboratories is a multi-mission laboratory managed and operated by National Technology \& Engineering Solutions of Sandia, LLC (NTESS), a wholly owned subsidiary of Honeywell International Inc., for the U.S. Department of Energy’s National Nuclear Security Administration (DOE/NNSA) under contract DE-NA0003525.
This written work is authored by an employee of NTESS. 
The employee, not NTESS, owns the right, title and interest in and to the written work and is responsible for its contents.
Any subjective views or opinions that might be expressed in the written work do not necessarily represent the views of the U.S. Government.
The publisher acknowledges that the U.S. Government retains a non-exclusive, paid-up, irrevocable, world-wide license to publish or reproduce the published form of this written work or allow others to do so, for U.S. Government purposes.
The DOE will provide public access to results of federally sponsored research in accordance with the DOE Public Access Plan.

AI Disclosure: During the preparation of this work the authors used Sandia National Laboratories' SandiaAI Chat and Claude.ai in order to correct spelling and grammar and reduce the length of the manuscript.
After using these tools, the authors reviewed and edited the content as needed and take full responsibility for the content of the publication. 

\bibliographystyle{plain}
\bibliography{references}
\end{document}